\documentclass{article}

\usepackage{multirow}

\usepackage{arxiv}
\usepackage[utf8]{inputenc} 
\usepackage[T1]{fontenc}    
\usepackage{hyperref}       
\usepackage{url}            
\usepackage{booktabs}       
\usepackage{amsfonts}       
\usepackage{nicefrac}       
\usepackage{microtype}      
\usepackage{lipsum}		
\usepackage{graphicx}
\usepackage{natbib}
\usepackage{doi}
\usepackage[inline]{enumitem}
\usepackage{multirow}
\usepackage{amssymb}
\usepackage{ulem}
\usepackage{amsmath}
\usepackage{soul}
\usepackage{multirow}

\newcommand{\modification}[1]{{{#1}}}
\newcommand{\changetext}[2]{#2}

\def\ie{\emph{i.e}, \space}

\title{ERA: Entity–Relationship Aware Video Summarization with Wasserstein GAN}

\date{} 					

\author{ \href{https://orcid.org/0000-0002-9244-173X}{\includegraphics[scale=0.06]{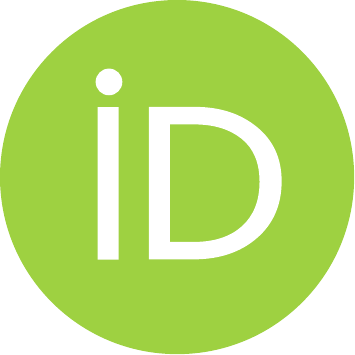}\hspace{1mm}Guande Wu}\\
	Tandon School of Engineering\\
	New York University\\
	Brooklyn, NY 11201 \\
	\texttt{guandewu@nyu.edu} \\
	\And
	\href{https://orcid.org/0000-0002-1559-1048}{\includegraphics[scale=0.06]{orcid.pdf}\hspace{1mm}Jianzhe Lin} \\
	Tandon School of Engineering\\
	New York University\\
	Brooklyn, NY 11201 \\
	\texttt{jianzhelin@nyu.edu} \\
	
	\And
	\href{https://orcid.org/0000-0003-2452-2295}{\includegraphics[scale=0.06]{orcid.pdf}\hspace{1mm}Claudio T. Silva} \\
	Tandon School of Engineering\\
	New York University\\
	Brooklyn, NY 11201 \\
	\texttt{csilva@nyu.edu} \\
}

\renewcommand{\shorttitle}{ERA: Entity–Relationship Aware Video Summarization with Wasserstein GAN}

\hypersetup{
pdftitle={ERA Preprint},
pdfsubject={CS.CV},
pdfauthor={Guande Wu},
pdfkeywords={Video Summarization, Graph Neural Network},
}

\begin{document}
\maketitle

\begin{abstract}
	Video summarization aims to simplify large-scale video browsing by generating concise, short summaries that diver from but well represent the original video. Due to the scarcity of video annotations, recent progress for video summarization concentrates on unsupervised methods, among which the GAN-based methods are most prevalent. This type of methods includes a summarizer and a discriminator. The summarized video from the summarizer will be assumed as the final output, only if the video reconstructed from this summary cannot be discriminated from the original one by the discriminator. The primary problems of this GAN-based methods are two-folds. First, the summarized video in this way is a subset of original video with low redundancy and contains high priority events/entities. This summarization criterion is not enough. Second, the training of the GAN framework is not stable. This paper proposes a novel Entity–relationship Aware video summarization method (ERA) to address the above problems. To be more specific, we introduce an Adversarial Spatio-Temporal network to construct the relationship among entities, which we think should also be given high priority in the summarization. The GAN training problem is solved by introducing the Wasserstein GAN and two newly proposed video-patch/score-sum losses. In addition, the score-sum loss can also relieve the model sensitivity to the varying video lengths, which is an inherent problem for most current video analysis tasks. Our method substantially lifts the performance on the target benchmark datasets and exceeds the current leaderboard Rank-1 state-of-the-arts (2.1\% F1-score increase on TVSum and 3.1\% F1-score increase on SumMe). We hope our straightforward yet effective approach will shed some light on the future research of unsupervised video summarization. The code is available online\footnote{https://github.com/jnzs1836/ERA-VSum}.

\end{abstract}


\section{Introduction}
\label{sec:intro}
As a primary source of recording information, video data on social networks are becoming the dominating form of information exchange. However, with the explosion of video data on different platforms (Youtube, Instagram, etc.), processing (cataloging, captioning, searching, etc.) these large number of videos manually according to their categories and subject matter would be frustrating and unintelligent. Therefore, the storage and compression had attracted researchers' attention. How to efficiently keep and browse these videos needed a rethinking. One plausible solution was summarizing the long video into a concise synopsis with the most salient and representative content. Such a synopsis could be hyper-lapse\cite{joshi2015real}, montage\cite{kang2006space, khosla2013large}, and storyboards\cite{gygli2014creating, gygli2015video}. In this study, we focused on the video storyboard. 

The existing method for obtaining a video storyboard was through video summarization. Because most of the accessible videos online were with no annotations, and it would be time-consuming to obtain these annotations through human labeling, the unsupervised video summarization (UVS) model was more practical. A most famous UVS model was realized by adversarial networks \cite{mahasseni2017unsupervised}. The notion of the framework was that the summary generator was trained to fool a discriminator, which tried its best to distinguish the features reconstructed from the summary. Many follow-up works were proposed in recent years, but two significant problems still existed and were ignored by researchers, from the perspectives of both the loss criterion and the model training.

\begin{figure*}
\begin{center}
\includegraphics[width=\textwidth]{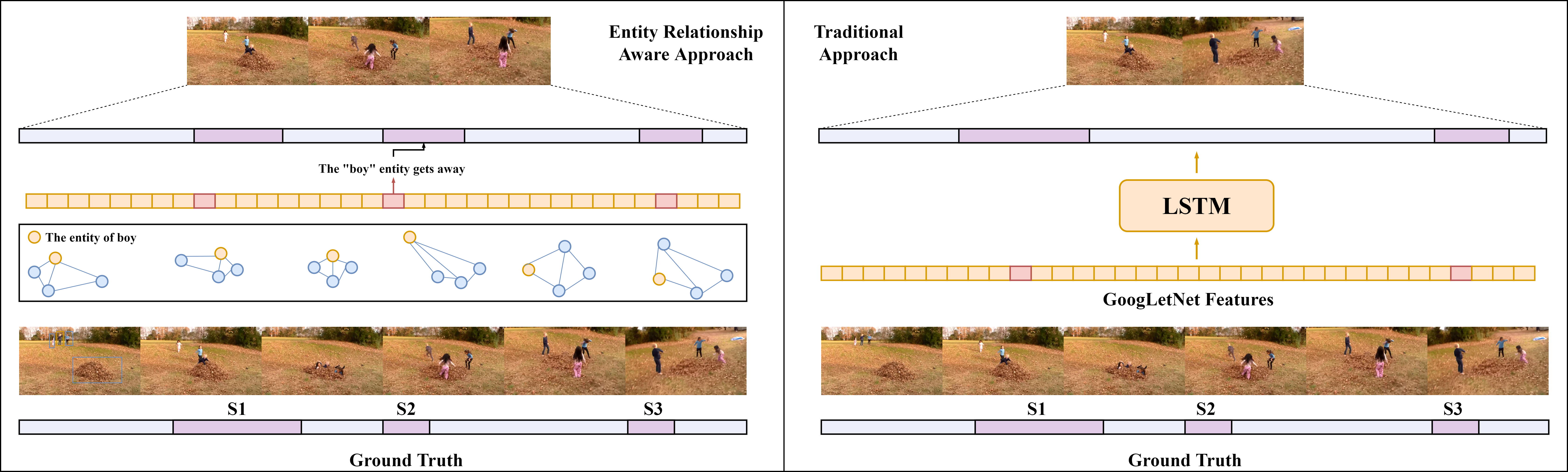}
\end{center}
   \caption{ 
   The clip of first 20 seconds of Kid's playing video in SumMe. The ground-truth summary has three shots in the period and we denote them as $S1, S2, S3$. When the traditional method can capture $S1$ and $S3$, it fall shorts in capturing $S2$. $S2$ describes that the body runs away from the leave stack because he got "attacked" by other kids. The features extracted from GoogLeNet may fail to capture the boy's movement. By comparison, the Spatio-Temporal Graph captures the change of the boy's relative position.
   }
\label{fig:motivation}
\end{figure*}

The criterion of defining model loss in existing models to generate the summarized video is to find frames which: a). contains the entities and events with high priority from the video. b). are with low repetition and redundancy. However, such a criterion might not be feasible to summarize practical scenarios when a trivial accident happens. For example, two kids collide with each other cannot be assumed as a key event but should be an essential accident. In this paper, we assume all these accidents to be associated with entity-relationship. Only when entities interact with each other will the accident happens. \changetext{Therefore, an entity-relationship aware video summarization method is proposed. The relationship of different entities is modeled by a novel Spatio-Temporal network, and changes of relationship will be easily captured and extracted in this way.}{Therefore, we propose an entity-relationship aware video summarization method. The relationship of different entities is modeled by a novel Spatio-Temporal network, and changes of relationship will be easily captured and extracted in this way.} In addition, different from existing methods, we introduce a novel feature extraction module to extract the scene context to help with the construction of entity-relationship.  A more detailed comparison of the proposed criterion and the traditional ones can be found in Fig. \ref{fig:motivation}.

The model training of the adversarial network based method is another problem. The discriminator in this type of model is not stable, while the existing methods rarely consider this. To deal with this problem, We discern two significant issues of the discriminator. Firstly, the BCE loss used in the discriminator can evoke extra training difficulty as it suffers from the vanishing gradient when there is little overlap between the generated and original samples. To solve this BCE loss problem, we instead use the earth moving distance in Wasserstein GAN \cite{arjovsky2017wasserstein} to formulate the loss. Another issue is the varying video length. The sparsity of feedback from the discriminator varies, which will mislead the generator. To deal with this problem, we introduce a novel patch mechanism to monitor this sparsity.

The rest of this paper is organized as follows. The related work is presented in section II, which is followed
by our proposed ERA model as in section III. We present our experimental results in section IV and give a conclusion in section V.

\section{Related Works}
\label{sec:related}
In this section, we will first generally review the existing unsupervised video summarization (UVS) models. Then we will introduce the Spatio-Temporal Graph network which is our baseline model.
\subsection{Unsupervised Video Summarization}
Unsupervised video summarization methods learn a video summary with the absence of the ground-truth labels. 
Earlier works explored various heuristics representing the frame importance and guiding the summarization. 
Ngo et al. proposed to summarize the video based on video structures and video highlights\cite{ngo2003automatic}. However, this method relied heavily on prior knowledge, which was not realistic in real-life scenarios.
Gygli \cite{gygli2014creating} et al. employed a segment-level visual interesteness score and selected the optimal subset of the segments based on the scores. This work was further extended to multi-objective based optimal subset selection as in \cite{gygli2015video}. However, the feature extraction and representation parts of these works were still weak. Recent work introduced the deep learning (DL) module for the representation of input videos \cite{rochan2018video,zhao2018hsa,wei2018video,rochan2019video}. A most representative branch of DL-based UVS methods was based on a generative adversarial manner, which replaced the human-defined heuristics with a learned discriminator \cite{apostolidis2019stepwise, kanafani2021unsupervised, mahasseni2017unsupervised, yuan2019cycle, jung2019discriminative}. This type of method included a feature generator (summarizer) and a discriminator. The summarized video from the generator would be assumed as the final output, only if the video reconstructed from this summary could not be discriminated from the original one by the discriminator. The earliest attempt of this type of method was made by Mahasseni et al., who introduced the LSTM based GAN for UVS for the first time \cite{mahasseni2017unsupervised}. Following it, CSNet added the local chunks and global stride (view) to the input features. These features enhanced the generator. Another similar work can be found in \cite{zhou2018deep} and \cite{ji2019video}, in which multiple features/attentional features were introduced to improve the performance of the feature generator. 
Unlike the existing feature generator, we merged the object-level and scene-level features on the generator side in our work. 
The idea of merging different sources of features had been explored by Kanafani et al. \cite{kanafani2021unsupervised} However, their approach only considered the visual features extracted from two vision models pre-trained on ImageNet. By comparison, we constructed a spatiotemporal graph of the detected objects and extracted the object-level features. 
\changetext{}{Park et al. \mbox{\cite{park2020sumgraph}} also exploit graph-based approach for video summarization. However, their work focuses on the relationships between the frames without the object-level relationships.}
Also, we introduce a novel discriminator, which was a rarely touched area in former works. Our work differed from the previous studies by replacing the discriminator with a critic used in Wassertein GAN and introducing a video-patch mechanism. 

\subsection{Spatio-Temporal Graph for the Video}
A key characteristic of video data is the associated spatial and temporal semantics \cite{liu2020learned, 971195}.
Spatiotemporal graph, which models the characteristics of objects and their relationships by a graph structure, can be to learn this spatio-temporal correlation \cite{brendel2011learning,tian2013spatiotemporal,geng2019spatiotemporal}.
The sptiotemporal graph at the very beginning was used to learn human activity\cite{brendel2011learning} and detect events \cite{chen2012efficient} in the videos. \changetext{}{An early attempt was also made on video summarization by Zhang et al. while their work relies on the hand-crafted features and does not utilize the deep learning approach\mbox{\cite{zhang2016context}}.}
Recently, spatiotemporal graph models had been extended to more general video processing applications \changetext{}{with deep learning}. For examples,
Wang et al. introduced Graph Convolution Network to process the spatiotemporal graph and perform action recognition\cite{ferrari_videos_2018}; Yan et al. modeled human body joints as a spatiotemporal graph and performed pose estimation based on a spatiotemporal Graph Convolution Network\cite{yan_spatial_2018}.
Other applications included action/object localization\cite{ghosh2020stacked, mavroudi2019neural,wang2021weakly}, video captioning\cite{pan2020spatio}, human re-identification \cite{liu2021spatial}, and gaze prediction\cite{fan2019understanding}, etc. In our work, we introduce the \changetext{}{deep} spatiotemporal graph to UVS task for the first time, and it is used to correlate the object-level features. Our work mainly follows Wang's work \cite{ferrari_videos_2018}. 
However, the object-level features can be noisy and sometimes unavailable in video summarization, making the spatiotemporal graph far from enough to capture all the cues in a particular video. Thus, besides the object-level features, we also complement the spatiotemporal graph with scene features, and make the prediction based on these two features.


\section{Method}
\begin{figure*}
\begin{center}
\includegraphics[width=0.95\textwidth]{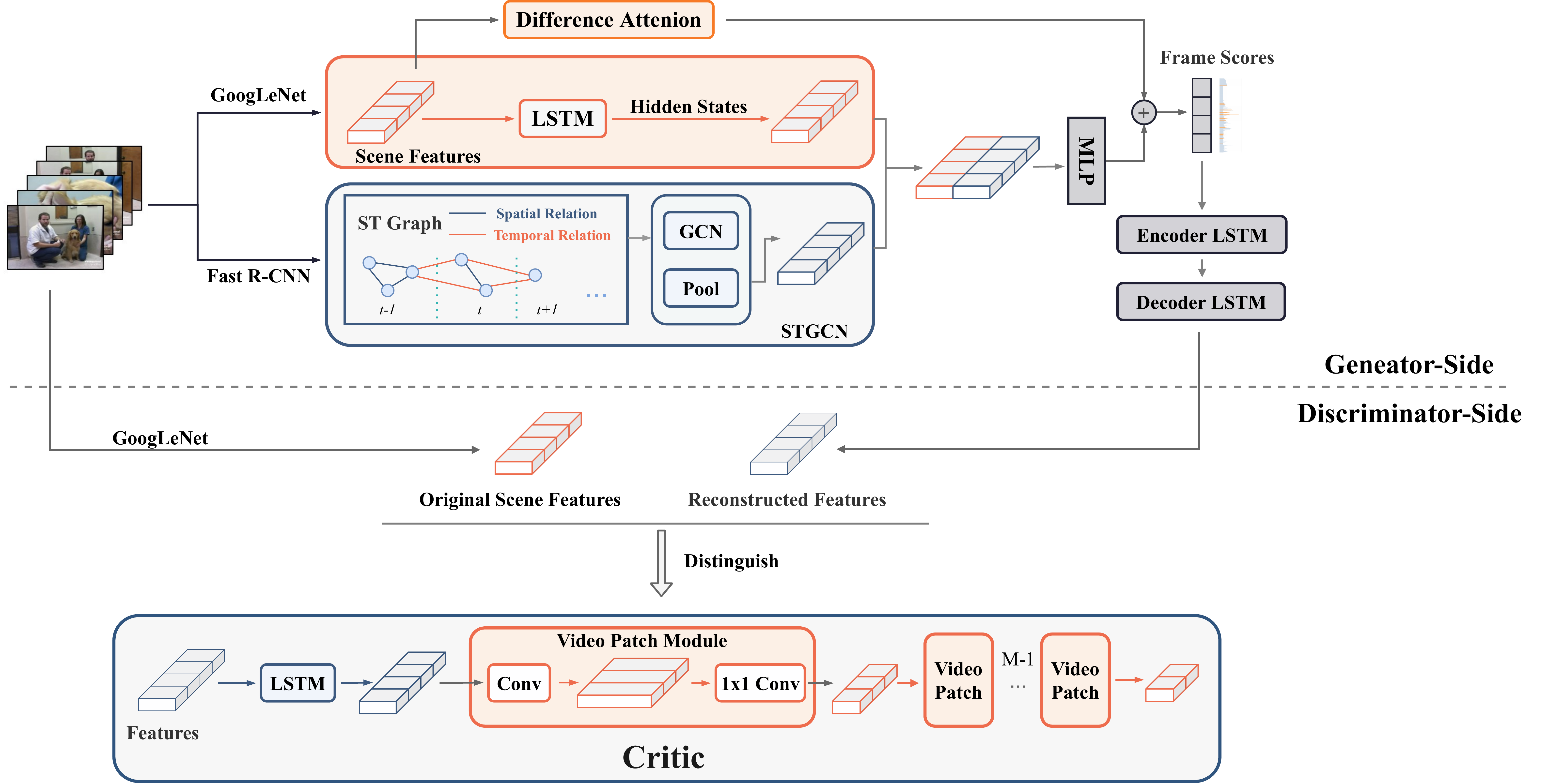}
\end{center}
   \caption{ 
   Our proposed methods can be categorized into generator-side and discriminator-side. The generator-side model combines the STGCN, LSTM and Difference Attention to predict the frame scores before the frame scores are exploited to reconstruct the video features by an encoder-decoder structure. The discriminator exploits the critic model proposed by W-GAN\cite{arjovsky2017wasserstein} and introduces a video patch module to distinguish the original and reconstructed features at a patch level.
      }

\label{fig:short}
\end{figure*}
\label{sec:method}
\changetext{In this section, we will describe our proposed methods for unsupervised video summarization}{In this section, we will describe our proposed ERA framework.}
We base our methods upon the adversarial unsupervised framework proposed by Mahasseni et al.\mbox{\cite{mahasseni2017unsupervised}} It consists a generator of VAE for reconstructing the summary-based visual features and a discriminator of LSTM. The VAE further comprises of three modules \ie{Summarizer, Encoder LSTM, Decoder LSTM}. 
\changetext{We first incorporate a Spatio-Temporal Network with a score-sum loss in the summarizer to model the relationships of different entities. }{We first propose a Spatio-Temporal Network-based summarizer with a score-sum loss to explicitly capture the entity relationships and realize ERA concept.}
Our Encoder LSTM and Decoder LSTM are identical to the original version.
Then to deal with the training difficulty, we replace the discriminator with the critic proposed in Wassertein GAN\mbox{\cite{arjovsky2017wasserstein}} and introduce a video-patch mechanism.

\subsection{Spatio-Temporal Graph Convolution Network (STGCN)}
To model the entities' relationships and capture their changes in the video, we incorporate a Spatio-Temporal Graph where each vertex represents an entity, and each edge models the entities' relationship.
We construct the graph by
\begin{enumerate*}
    \item extracting entities from each frame via Fast R-CNN;
    \item inferring the entities' relationships by a set of heuristics.
\end{enumerate*}
Then the graph is fed into the Graph Convolution Network (GCN) to learn a graph representation. 
Finally, we can obtain entity-relationship aware features by performing temporal pooling\cite{pan2020spatio} on the graph representation.
The entity-relationship aware features can be noisy in the particular video due to the limited object detection accuracy and the sparsity of entities. To address it, we complement the features with other sources of features in the score prediction. Specifically, we combine the visual features extracted by GoogLeNet similar to \cite{mahasseni2017unsupervised, apostolidis2019stepwise} and the difference attention proposed by \cite{jung2019discriminative}.

\subsubsection{Spatio-Temporal Graph}
Given a video of $T$ frames, we first run Fast R-CNN on each frame to extract the entities and their features. We represent the entity feature set by $\Omega = \{\mathbf{o_1^1, o_2^1, o_{N_1}^{1}, ..., o_i^t,...,o_{N_t}^t,..., o_{N_T}^T}\}$ where $o_{i}^t$ represents the feature vector of $i$-th entity in $t$ frame. Besides, $N_t$ denotes the number of entities in frame $t$.
Based on the extracted entities, we define a graph as
\begin{equation}
    G = (\Omega,E)
\end{equation}
where $E=\{w_{i,j}\}$ is a set of edges between the different entities. We can also represent $E$ as an adjacency matrix of the entities.
We specify the edge weights by the following spatial and temporal graphs.

\noindent \textbf{Spatial Graph}
\changetext{
The spatial graph models the spatial relationships between the entities within a frame. 
Since the intra-frame entities' relationships are closely correlated with the relative positions\cite{ferrari_videos_2018}, we represent the edge weight by the value of Intersection Over Unions(IOU) similar to the previous works \cite{pan2020spatio}.
}{The different entities can be related in the spatial domain. To model the intra-frame entities' relationships, we propose a spatial graph where the entities within the same frame are connected. Since the spatial relationships heavily depend on the spatial proximity, we weight the relationships by the value of Intersection Over Unions(IOU) similar to the previous works \mbox{\cite{pan2020spatio}}.
}

We denote the IOU between the entities $o^t_i$ and $o^t_j$ of the frame $t$ as $\sigma^t_{ij}$. Then we assign the edge weight by the normalized IOU as follows:
\begin{equation}
    G^{S^t}_{i,j} = \frac{exp(\sigma^t_{ij})}{\Sigma_{j=1}^{N_t} exp(\sigma^t_{ij})}
\end{equation}
where $G^{S^t}_{i,j}$ is the edge weight between the entities $o^t_i$ and $o^t_j$ in the frame $t$.

\noindent \textbf{Temporal Graph}
\changetext{The temporal graph models the chronological relationships,  between the entities in two adjacent frames. Following the previous works \mbox{\cite{pan2020spatio, ferrari_videos_2018}}, we represent the relationships by the cosine similarity of the entity features as follows:}
{The same entities can appear in different frames with changing positions, shapes and poses. To capture the inter-frame correlation, we construct a temporal graph where the entities in two adjacent frames are linked according to their feature similarity. Following the previous works \mbox{\cite{pan2020spatio, ferrari_videos_2018}}, we derive the edge weight by the cosine similarity of the entity features as follows:
}

\begin{equation}
    G^{T^t}_{i,j} = \frac{exp(cos(\mathbf{o_i^t, o_j^{t+1}}))} {\Sigma_{j=1}^{N_{t+1}} exp(cos(\mathbf{o_i^t, o_j^{t+1}}))}
\end{equation}
where $G^{T^t}_{i,j}$ is the edge weight between the entities $o^t_i$ and $o^{t+1}_j$.

\noindent \textbf{Spatio-Temporal Graph}
After obtaining the intra-frame spatial graph and inter-frame temporal graph, we combine them together to form a the adjacency matrix $E$.
\begin{equation}
   E = 
    \begin{bmatrix}
G^{s}_1 & G^{t}_{12} & 0 & \dots & 0 \\
0 & G^{s}_{2} & G^{t}_{23} & \dots & 0 \\
0 & 0 & G^{s}_{3} & \dots & 0 \\
\vdots & \vdots & \vdots & \ddots & 0 \\
0 & 0 &  0 & \dots & G^{s}_T
\end{bmatrix}	
\end{equation}

\subsubsection{Graph Convolution and Score Prediction}
Following \cite{ferrari_videos_2018}, we apply a graph convolution network on the Spatio-Temporal graph to learn a graph representation. 
\changetext{Then we perform a temporal pooling\cite{pan2020spatio} on the graph representation to extract the entity-relationship aware features of different frames as $F_o=\{\mathbf{f^1_o}, \mathbf{f^2_o},...\mathbf{f^t_o},..., \mathbf{f^T_o}\}$.}{Then we perform a temporal pooling\mbox{\cite{pan2020spatio}} on the obtained representation and extract the frame features as $F_o=\{\mathbf{f^1_o}, \mathbf{f^2_o},...\mathbf{f^t_o},..., \mathbf{f^T_o}\}$.}
 
Since the entity-relationship aware feature are noisy and unstable, we fusion it with another two sources of features.
The first source of features is the scene features extracted from the pool5 layer of GoogLeNet. We denote it as $F_s = \{\mathbf{f^0_{s}}, \mathbf{f^1_{s}},...,\mathbf{f^t_{s}}..., \mathbf{f^T_{s}}\}$. Then, we concatenate $F_s$ and $F_o$ into a merged feature set $F^*$, which will be fed into a MLP to predict frame scores $\mathbf{s^*} = [s^*_1, s^*_2, ...,s^*_T]^T$. Another source of features is the difference attention proposed by \cite{jung2019discriminative}. 
\changetext{We directly calculate the frame scores $\mathbf{s^d} = [s^d_1, s^d_2, ..., s^d_t,..., s^d_T]^T$
based on the attention module. Then, we can obtain the final frame score by taking the average the of two scores.}{We derive the final frame scores by take average of the frame scores $\mathbf{s^d} = [s^d_1, s^d_2, ..., s^d_t,..., s^d_T]^T$ calculated by the difference attention and the score $\mathbf{s^*}$}.

\subsection{Score-Sum Loss}
An training issue is that the summarizer tends to assign high scores to all the frames. The sparsity loss term partially address the issue in \cite{mahasseni2017unsupervised}. 
However, the loss term is calculated against a fixed summary rate $\sigma$ (in this case, $\sigma=15\%$), which is not always the case. 
Thus, we propose a score-sum loss to penalize the summarizer for assigning high scores as follows:
\begin{equation}
    L_{sum} = \frac{\Sigma s_t}{\sqrt{T}}
\end{equation}
where $s_t$ refers to the score of frame $t$. The loss is only used in training the summarizer.
\subsection{Wasserstein GAN with Video Patch Mechanism}
\subsubsection{Wasserstein GAN}
Though used widely, GANs are often criticized as difficult to train. One of the reasons stems from the vanishing gradient issue caused by the Jensen-Shannon (JS) divergence. 
To address the issue, Wasserstein GAN replaces the discriminator with a critic regularized by a gradient penalty\cite{arjovsky2017wasserstein}.
Following it, we also employ a critic to minimize the loss function below:
\begin{equation}
    L(c) = ||E(c(x)) - E(c(x'))||_2 + \lambda (||\nabla c||_2 - 1 )^2
\end{equation}
where $x$ is the original features and $x'$ is the reconstructed features. $c$ represents the critic function and $\nabla c$ represents $c$'s gradients. $\lambda$ is a hyper-parameter for the gradient penalty.

\subsubsection{Video Patch Mechanism}
Another training issue comes from the varying video lengths. The sparsity of feedback from the discriminator varies according to the different video lengths, which will mislead the generator. We address the issue by introducing a video-patch mechanism following the notion of PatchGAN\cite{isola2017image}.
Given a sequence of video features $H=\{h_1, h_2,...,h_t, ..., h_T\}$ with a feature size of $K$(either reconstructed or original), we employ a 1-D convolution network to reduce the sequence length and patch the frames. 
The convolution network consists of $M$  building blocks, each of which can reduce the sequence length to a fifth. 
The building block consists of two 1-D convolution layers. The first layer reduces the sequence length with a stride of five and double the feature dimension.
Then the second convolution layer performs $1\times 1$ convolution on the sequence to reduce the feature dimension to $K$. Thus, the building block can output a shorter sequence with same feature size as $H^{m+1} = \{h^m_1, h^m_2, ..., h^m_{[\frac{T}{5}]}\}$ where $m$ refers the $m$-th building block. After $M$ building blocks, we can obtain an aggregated sequence of hidden features, $H^M = \{h^M_1, h^M_2, ..., h^M_{[\frac{T}{5^M}]}\}$. Each element in the sequence can have a receptive field of $5^M$ and thus attend to a patch of $5^M$ frames in the video.

\section{Experiments}
\label{sec:experiments}

\subsection{Experiment Settings}
\noindent \textbf{Implementations \space}
Following the previous works\cite{mahasseni2017unsupervised, apostolidis2019stepwise}, we downsample the videos to 2 fps. 
We exploit Fast R-CNN provided by Detectron2  \cite{wu2019detectron2} to extract the entity-level features.
We employ a three-layer Graph Convolution Network with the shortcut connections between the layers \cite{he2016deep} to process the spatiotemporal graph. 
We train our model with Adam optimizers with a learning rate of 1e-4 and 0.1 times after ten epochs.

\noindent \textbf{Datasets and Evaluation Metric \space}
We evaluate our approach on two widely used benchmark datasets \ie{SumMe\cite{gygli2014creating} and TVSum\cite{song2015tvsum}}. 

We use the standard 5-fold cross-validation for both datasets. 
For a fair comparison, we first employ the randomly generated data splits available from \cite{apostolidis2019stepwise},
\changetext{}{which are also used by \mbox{\cite{apostolidis2020ac, apostolidis2020unsupervised, phaphuangwittayakul2021self, kanafani2021unsupervised}}}
 However, \cite{kanafani2021unsupervised} reports that a non-trivial number of the videos are not part of any test set of the five data splits. \changetext{In other words, the model's performance is never tested against particular videos.}{} Thus, we also generate non-overlapping splits where all the videos occur in the test splits precisely once.
We assess the result by the harmonic F-measure used in \changetext{\cite{zhang2016video}}{\mbox{\cite{zhang2016video,apostolidis2019stepwise, apostolidis2020ac, kanafani2021unsupervised}}}. 
\changetext{}{It compares the machine-generated summary with the multiple user-annotated summaries to compute a set of F-measures. By taking the average and maximum value of the F-measure set, we can derive two different F-measures (Avg and Max in Table-\mbox{\ref{tab:comparison}}) to evaluate the machine summary.}

\subsection{Quantitative Analysis}
\begin{table}
\begin{center}
\begin{tabular}{|c|c|c|c|c|c|c|c|}
\hline
\multirow{2}{*}{Dataset}                                      & \multirow{2}{*}{Method}       & \multicolumn{2}{c|}{F1}                           & \multicolumn{2}{c|}{F1'}                              & \multicolumn{2}{c|}{F1*}                              \\ \cline{3-8}
 &  & Avg     & Max                   & Avg                       & Max                       & Avg                       & Max                       \\ \hline
\multirow{5}{*}{SumMe} & $\mathrm{SUM-Ind_{LU}}$ \cite{yaliniz2019unsupervised}      & -                         & 51.9                  & 22.1                      & 46.0                      &           19.1 & 42.5                      \\ \cline{2-8} 
& CSNet  \cite{jung2019discriminative}      & - & 51.3  & 22.7  & 48.1 & 19.0 & {43.2} \\ \cline{2-8} 
  & SUM-GAN-AAE \cite{apostolidis2020unsupervised} &-& 48.9& 22.8 & 47.1& \underline{19.2} & 41.5                      \\ \cline{2-8} 
    & MCSF   \cite{kanafani2021unsupervised}      & -                         & 46.0                   & 21.0                      & 46.0                      & 17.4                          & 41.7                      \\ \cline{2-8}
     \multicolumn{1}{|l|}{}  &   \modification{DSR-RL-GRU \mbox{\cite{phaphuangwittayakul2021self}}} & \modification{-}                         & \modification{50.3}                  & \modification{22.6}                   & \modification{\underline{50.3}}                    &           \modification{18.2} & \modification{40.2}                     \\ \cline{2-8} 
   \multicolumn{1}{|l|}{}                       & \modification{AC-SUM-GAN \mbox{ \cite{apostolidis2020ac}}}        & \multicolumn{1}{c|}{\modification{-}}     & \multicolumn{1}{c|}{\modification{{50.8}}} & \multicolumn{1}{l|}{{\modification{\underline{22.9}}}} & \multicolumn{1}{c|}{\modification{\textbf{50.8}}} & \multicolumn{1}{l|}{\modification{19.0}}     & \multicolumn{1}{l|}{\modification{\underline{43.9}}}    \\ \cline{2-8} 
    
    & ERA (Ours) & -                         & -                     & \textbf{23.2}             & {48.8}             & \textbf{19.3}                          & \textbf{46.3}             \\ \hline
\multicolumn{1}{|l|}{\multirow{5}{*}{TVSum}} & $\mathrm{SUM-Ind_{LU}}$ \cite{yaliniz2019unsupervised}     & \multicolumn{1}{l|}{61.5} & \multicolumn{1}{c|}{-} & \multicolumn{1}{l|}{58.7} & \multicolumn{1}{l|}{80.7} & \multicolumn{1}{l|}{56.6}     & \multicolumn{1}{l|}{{79.3}}     \\ \cline{2-8} 
\multicolumn{1}{|l|}{}                       & CSNet  \cite{jung2019discriminative}      & \multicolumn{1}{l|}{58.8} & \multicolumn{1}{c|}{-} & \multicolumn{1}{l|}{56.4} & \multicolumn{1}{l|}{77.7} & \multicolumn{1}{l|}{54.4}     & \multicolumn{1}{l|}{77.4}     \\ \cline{2-8} 
\multicolumn{1}{|l|}{}    & SUM-GAN-AAE\cite{apostolidis2020unsupervised}  & \multicolumn{1}{l|}{58.3} &  \multicolumn{1}{c|}{-} & \multicolumn{1}{l|}{57.7} & \multicolumn{1}{l|}{\textbf{81.6}} & \multicolumn{1}{l|}{55.2}     & \multicolumn{1}{l|}{77.9}     \\ \cline{2-8} 
\multicolumn{1}{|l|}{}                       & MCSF \cite{kanafani2021unsupervised}         & \multicolumn{1}{c|}{59.1}     & \multicolumn{1}{c|}{-} & \multicolumn{1}{l|}{59.1} & \multicolumn{1}{l|}{81.2} & \multicolumn{1}{l|}{\underline{58.3}}     & \multicolumn{1}{l|}{78.1}     \\ \cline{2-8} 

\multicolumn{1}{|l|}{}  &  \modification{DSR-RL-GRU \mbox{\cite{phaphuangwittayakul2021self}}}     & \multicolumn{1}{l|}{\modification{60.2}} & \multicolumn{1}{c|}{\modification{-}} & \multicolumn{1}{l|}{\modification{\underline{60.2}}} & \multicolumn{1}{l|}{\modification{81.3}} & \multicolumn{1}{l|}{\modification{\underline{58.3}}}     & \multicolumn{1}{l|}{{\modification{79.3}}}     \\ \cline{2-8} 
\multicolumn{1}{|l|}{}                       &  \modification{AC-SUM-GAN \mbox{\cite{apostolidis2020ac}}}         & \multicolumn{1}{c|}{\modification{60.6}}     & \multicolumn{1}{c|}{\modification{-}} & \multicolumn{1}{l|}{{\modification{\textbf{60.6}}}} & \multicolumn{1}{c|}{\modification{81.2}} & \multicolumn{1}{l|}{\modification{57.5}}     & \multicolumn{1}{l|}{\modification{\underline{80.2}}}    \\ \cline{2-8} 
\multicolumn{1}{|l|}{}                       & ERA (Ours) & \multicolumn{1}{c|}{-}     & \multicolumn{1}{c|}{-} & \multicolumn{1}{l|}{58.0}     & \multicolumn{1}{l|}{\underline{81.5}}     & \multicolumn{1}{l|}{\textbf{58.9}} & \multicolumn{1}{l|}{\textbf{81.4}} \\ \hline
\end{tabular}
\end{center}
\caption{Comparison with the state-of-the-art unsupervised approaches: The experiments are conducted with the splits of  \cite{apostolidis2019stepwise}($F_1'$) and non-overlapping splits ($F_1^*$). $F_1$ refers the reported results in the corresponding papers.
\label{tab:comparison}
}
\end{table}
 
We compare our method with six state-of-the-art unsupervised methods \ie{$\mathrm{SUM-Ind_{LU}}$\cite{yaliniz2019unsupervised}, CSNet\cite{jung2019discriminative}, SUM-GAN-AAE\cite{apostolidis2020unsupervised} and MCSF\cite{kanafani2021unsupervised}}, \changetext{}{DSR-RL-GRU\mbox{\cite{phaphuangwittayakul2021self}} and AC-SUM-GAN\mbox{\cite{apostolidis2020ac}}}.
The official implementations for MCSF\footnote{https://gitlab.uni-hannover.de/hussainkanafani/unsupervised-video-summarization}
, SUM-GAN-AAE\footnote{https://github.com/e-apostolidis/SUM-GAN-AAE}, DSR-RL-GRU\footnote{https://github.com/phaphuang/DSR-RL} and AC-SUM-GAN\footnote{https://github.com/e-apostolidis/AC-SUM-GAN} are available online. We exploit the unofficial implementations of $\mathrm{SUM-Ind_{LU}}$ provided by \cite{kanafani2021unsupervised} and verify its identity to the original paper. Then, we reimplement the CSNet since the authors did not provide their source code. We compare the different approaches based on two versions of the data splits described above. For the overlapping splits, we quote the experiment results of MCSF, SUM-GAN-AAE, and $\mathrm{SUM-Ind_{LU}}$ from \cite{kanafani2021unsupervised}. 
\changetext{}{\mbox{\cite{phaphuangwittayakul2021self, apostolidis2020ac}} also employ the same splits but they only present the average F-measure for TVSum and maximum F-measure for SumMe. Thus, we cite these available F-measures directly and rerun their official implementations to get the rest F-measures.
}
It can be observed that our approach outperforms all the competitors on the non-overlapping splits of the two datasets. Furthermore, we observe that the improvement on TVSum (0.6\%) is not as significant as it on SumMe(3.1\%). An explanation can be that TVSum videos contain more discontinuous scenes, hindering the temporal relationships between the objects in different scenes.

\subsection{Ablation Study}
\changetext{The success of ERA attributes to the various approaches we propose. }{}
Our proposed approaches can be generally divided into two categories, \ie{generator-side and discriminator-side}. 
To analyze the effects of them, we conduct two individual ablation studies for the two sides correspondingly. 
We also adopt SumMe as our ablation study dataset similar to \cite{jung2019discriminative}.

\subsubsection{Ablation Study for the Generator-Side Approaches}
Our generator-side approaches include the Spatio-Temporal Network and the score-sum loss. 
However, since our model also incorporates the difference attention proposed by \cite{jung2019discriminative}, it is necessary to analyze the usability and necessarity of the mechanism. Thus, we conduct the ablation study by adding the three approaches step by step and studying their effects. Results are provided in Table \ref{tab:abl1}, from which we can obtain the following key findings.


\noindent \textbf{Solely using an approach can be ineffective.} Exp. 1, 2, 3 show that the models with the sole approach can not even outperform the baselines. An explanation stems from the noisy entity-aware features. Thus, it is not reliable to depend solely on entity-aware features.

\noindent \textbf{Difference attention is effective but not necessary.} Exp. 5 and 7, show that difference attention boosts the performance. However, the comparison between Exp. 5 and other experiments, also prove that the model can achieve promising performance without the module. 

    


\subsubsection{Ablation Study for the Discriminator-Side Approaches}
\begin{table}[]
\begin{center}
\begin{tabular}{ccccc}
Exp.  & STGCN & Diff & SSum & F1    \\ \hline
0 &     &      &      & 39.36 \\ \hline
1 &   \checkmark   &      &      & 38.97 \\ \hline
2 &       &   \checkmark    &      & 36.94 \\ \hline
3 &       &      &    \checkmark   & 39.35 \\ \hline
4 &    \checkmark    &   \checkmark    &      & 39.56 \\ \hline
5 &    \checkmark    &      &   \checkmark    & 43.17 \\ \hline
6 &       &    \checkmark   &   \checkmark    & 39.65 \\ \hline
7 &    \checkmark    &    \checkmark   &    \checkmark   & \textbf{46.25} \\ \hline
\end{tabular}
\end{center}
\caption{Ablation Study for Generator-Side Approaches}
\label{tab:abl1}
\end{table}

\begin{table}[]
\begin{center}

\begin{tabular}{ccc}
Model             & SUM-GAN & STGCN \\ \hline
GAN               & 39.22   & 41.23 \\ \hline
WGAN              & \textbf{40.55}   & 42.66 \\ \hline
WGAN + Patch Loss & 39.74   & \textbf{46.25}
\end{tabular}
\end{center}
\caption{Ablation study Discriminator-Side Approaches}
\label{tab:abl2}
\end{table}

Our discriminator-side improvement involves the W-GAN framework and video-patch mechanism. To verify the effectiveness of them, we train our model and the baseline $SUM-GAN$ by incorporating different discriminators \ie{vanilla discriminator, critic of W-GAN and critic with patch mechanism}.
The experiment result is delivered in Table \ref{tab:abl2}.
From the experiments on the baseline, we only observe minor improvement made by our proposed W-GAN and video-patch mechanism. However, we observe a steadier and quicker training process when using them. 
Based on the experiments on ERA, we find using both W-GAN and patch mechanism can help our model achieve much better results. We think the difference can be caused by the noisy entity-aware features. Since the used features are noise, ERA is more likely to generate the low-quality summary in the early training stage. The reconstructed features based on the low-quality summary can have minor support with the original features. Thus, the vanishing gradiant problem of JS Divergence haunts ERA's training progress. By contrast, W-GAN addresses the issue and can promote the performance of ERA.

\begin{figure}
\begin{center}
    \begin{tabular}{cc}
  \includegraphics[width=0.3 \textwidth]{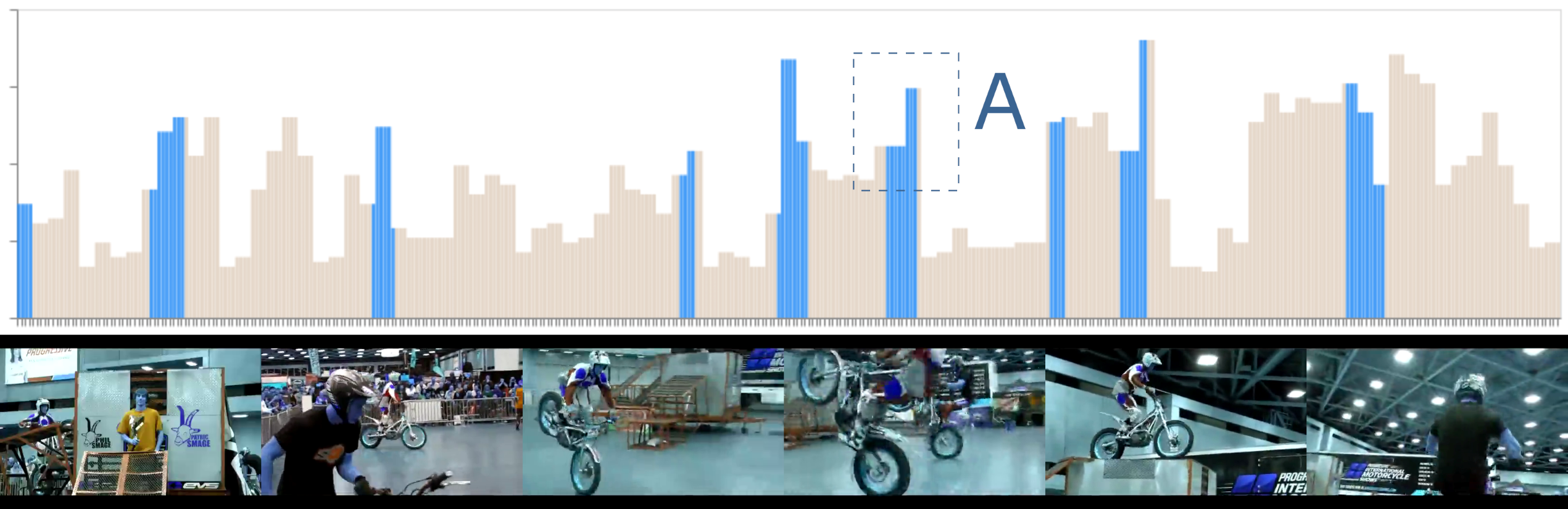} &   \includegraphics[width=0.3 \textwidth]{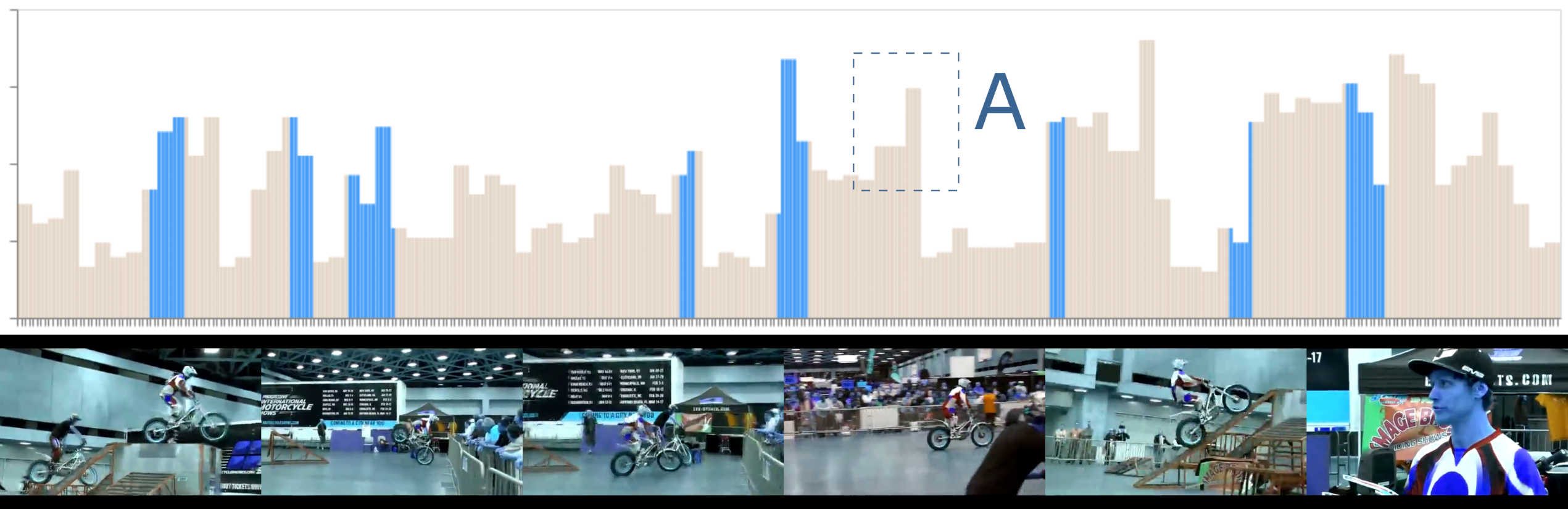} \\
(a) ERA (Ours) & (b) MCSF\cite{kanafani2021unsupervised} \\
 \includegraphics[width=0.3 \textwidth]{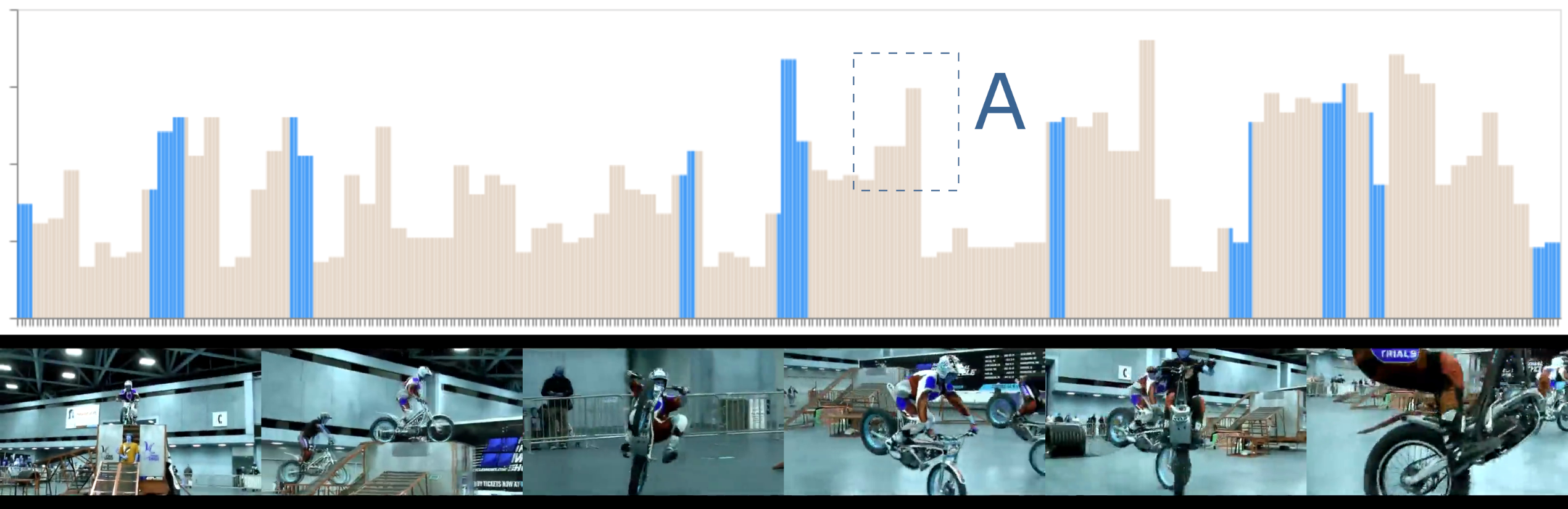} &   \includegraphics[width=0.3 \textwidth]{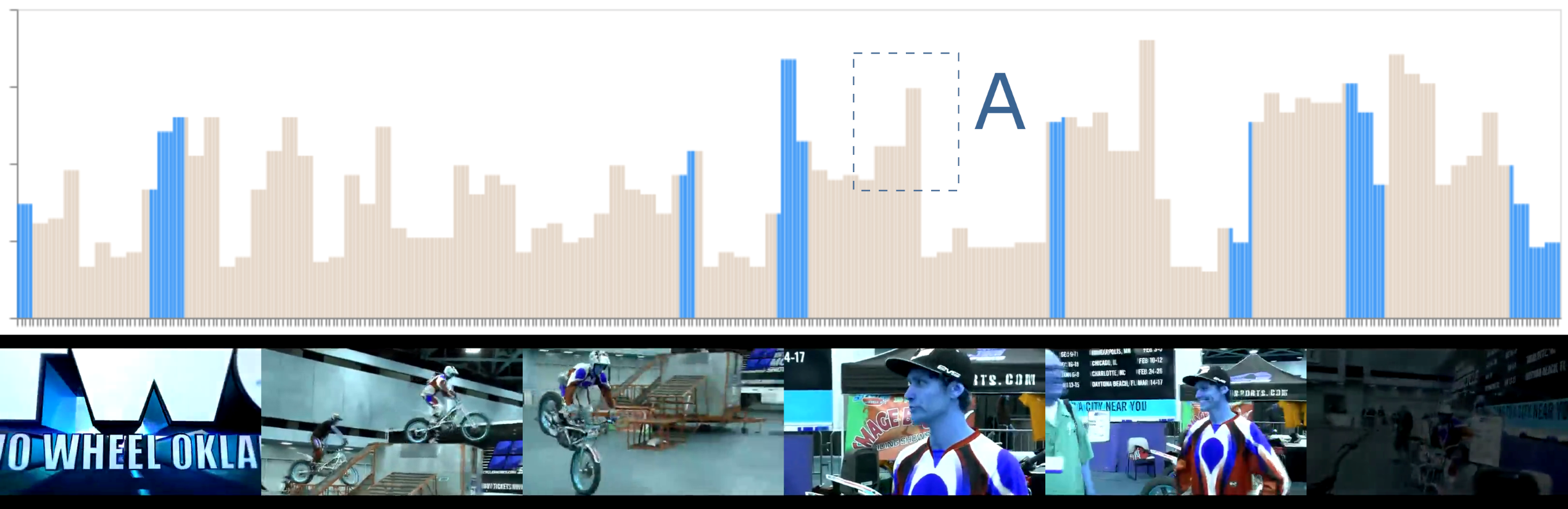} \\
(c) SUM-GAN-AAE \cite{apostolidis2020unsupervised} & (d) $\mathrm{SUM-Ind_{LU}}$\cite{yaliniz2019unsupervised} \\

\end{tabular}

\end{center}
\caption{Qualitative Analysis on TVSum Video-45}
\label{fig:qualitative}
\end{figure}

\subsection{Qualitative Analysis}
To illustrate the selection patterns of our summarization model, we visualize the selected frames and the ground-truth frame scores of the Video-45 in TVSum, shown in Figure \ref{fig:qualitative}.
Our method covers the peaks of the video, confirming it can capture the key-shots of the videos.
For example, our method is the only one to capture the peak $A$ in the video.

\section{Conclusion}
\label{sec:conclusion}
In this work, we study unsupervised video summarization (UVS) with adversarial learning. A novel Entity–Relationship Aware (ERA) video summarization is proposed in this paper. The method is made up of two parts, the generator and the discriminator. For the generator, we propose a novel Spatio-Temporal Graph Convolutional Network to model the entity-level features. For the discriminator, we employ Wasserstein GAN and propose a patch mechanism to deal with the varying video length. The effectiveness of the proposed ERA is verified on the TVSum and SumMe datasets.

\bibliographystyle{unsrtnat}
\bibliography{arxiv}  






\end{document}